\documentclass{article}
\usepackage{ijcai24}

\usepackage{times}
\usepackage{soul}
\usepackage{url}
\usepackage[hidelinks]{hyperref}
\usepackage[utf8]{inputenc}
\usepackage[small]{caption}
\usepackage{graphicx}
\usepackage{amsmath}
\usepackage{amsthm}
\usepackage{booktabs}
\usepackage{algorithm}
\usepackage{algorithmic}
\usepackage[switch]{lineno}
\usepackage{amsfonts}
\usepackage{siunitx} 
\usepackage{mdframed}
\usepackage{adjustbox}

\title{Aligning Large Language Models with Counterfactual DPO}

\author{
    Bradley Butcher
    \affiliations
    Independent Researcher
    \emails
    braddpbutcher@gmail.com
}

\begin{document}

\maketitle

\begin{abstract}
Advancements in large language models (LLMs) have demonstrated remarkable capabilities across a diverse range of applications. These models excel in generating text completions that are contextually coherent and cover an extensive array of subjects. However, the vast datasets required for their training make aligning response styles during the pretraining and instruction tuning phases challenging. Consequently, an additional alignment phase is typically employed, wherein the model is further trained with human preference data to better align its outputs with human expectations. While this process doesn't introduce new capabilities per se, it does accentuate generation styles innate to the model. This paper explores the utilization of counterfactual prompting within the framework of Direct Preference Optimization (DPO) to align the model's style without relying on human intervention. We demonstrate that this method effectively instils desirable behaviour, mitigates undesirable ones, and encourages the model to disregard inappropriate instructions. Our findings suggest that counterfactual prompting with DPO presents a low-resource way to fine-tune LLMs to meet the demands for responsible and ethically aligned AI systems.

\end{abstract}
\section{Introduction}

Recent progress in LLMs has elevated their text generation capabilities to a commercially viable standard~\cite{Qin2023Is,Vilar2022Prompting,Chowdhery2022PaLM}. These models are increasingly employed across diverse sectors, including chatbots, customer support assistants, retrieval-augmented generation (RAG), and structured data extraction ~\cite{soni2023large,lu2021text2event,gao2023retrieval}. As their integration becomes more widespread in professional and personal settings, aligning these models with user preferences has become crucial.

The training of LLMs typically involves three stages ~\cite{ziegler2019fine}. Initially, in the general pretraining stage, the model learns the structure of the language, gains context understanding, and develops the ability to generate coherent and relevant text~\cite{radford2018improving}. This is achieved by training the LLM to predict the next word in a sequence from a large corpus, including books, websites, and other texts, an approach known as causal language modelling (CLM)~\cite{vaswani2017attention}.

Following this, the model undergoes instruction fine-tuning, employing the same CLM objective but with data structured as pairs of instructions and responses. This phase enables the model to learn specific response patterns to various instructions, building on the foundational knowledge from the pretraining phase.

The final stage, alignment with human preferences, aims to refine the model's output for specific contexts, such as reducing hate speech or offering more concise responses~\cite{ziegler2019fine}. This typically involves human annotation, where users select their preferred response from multiple options generated by the LLM. However, this process is inherently limited, as the model must already be capable of generating both styles of response; the process merely emphasizes one style over the other. Furthermore, this approach is constrained by the need for extensive human annotations and the untargeted nature of preference selection.

Addressing these limitations, this paper introduces a novel approach using counterfactual prompts in the alignment stage, eliminating the need for human intervention. Our method utilizes counterfactual styles within the Direct Preference Optimization (DPO) framework to guide the model's output~\cite{rafailov2023direct}. This approach simplifies the alignment process and enhances scalability by directly injecting desired stylistic preferences into the model, a significant departure from traditional Reinforcement Learning from Human Feedback (RLHF) methods that depend on the availability and selection of styles by human evaluators. Our experiments are aimed at evaluating how effectively this method encourages the use of desired latent styles, minimizes the use of unwanted styles, and ignores inappropriate instructions. The results demonstrate that through counterfactual prompts combined with DPO, our approach can effectively reduce hallucinations, decrease bias, and ignore adversarial instructions. Our method offers a more controlled and efficient way to shape LLM responses, whilst maintaining the performance of the original model. 

With new global policies and rules for LLMs coming into play, it's important to make sure these models meet set behavior standards before they're released~\cite{europarl_2023_aiact}. In today's world, where open-source models are common, simply using prompts to guide behavior is not enough. Our method provides a way to deal with this issue. It allows for the embedding of certain behaviors in models before they are made available, offering researchers a practical tool to ensure models act as intended from the start.

\begin{figure*}[ht] 
    \centering
    \includegraphics[width=\textwidth]{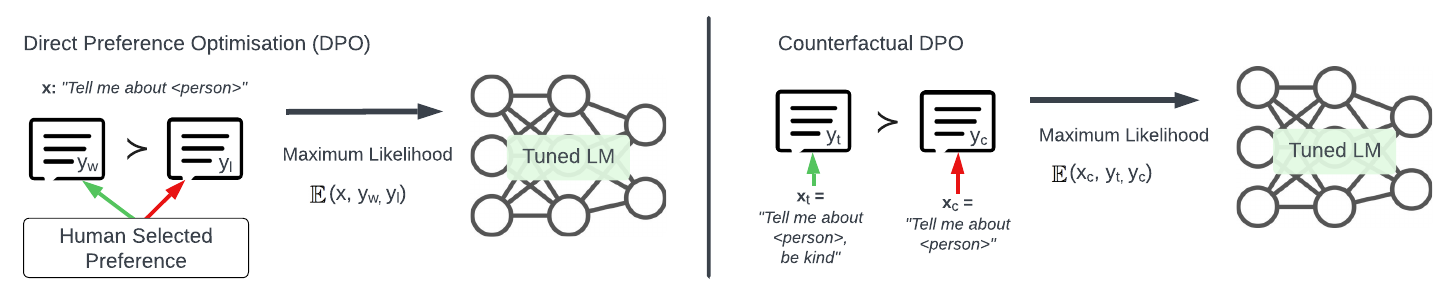} 
    \caption{An illustration of DPO and Counterfactual DPO (ENC). On the left, DPO is as normal, where human preference information is used to fine-tune the LLM policy via maximum likelihood. On the right, counterfactual DPO is used, desired style information is used to generate treatment and control prompts and responses. The treatment is assumed to be the preferred generation. DPO proceeds as normal, assuming the control prompt was used to generate both responses. We include diagrams of the other method configurations in supplementary material.} 
    \label{fig:figure1} 
\end{figure*}

\section{Background}

In the RLHF paradigm, LLMs are seen as policy models, with token selection as actions. This policy is optimized to maximize its token selection per human preference~\cite{ziegler2019fine}. It is generally applied following an instruction tuning phase, where a pre-trained language model is fine-tuned on curated data for tasks such as summarization, instruction following, or dialogue, resulting in a model $\pi_{sft}$. It is generally formulated as three steps:

\begin{itemize}
    \item \textbf{Data Collection:} The first step is to collect human preference data. Pairs of responses $(y_1, y_2)$ generated by $\pi_{sft}$ given a prompt $x$ are evaluated by human labelers to produce winning and losing responses $(x; y_w, y_l)$.
    \item \textbf{Reward Modelling Phase:} The collected preference data is used to model a reward function, commonly the Bradley-Terry model ~\cite{bradley1952rank,ziegler2019fine}, leading to a parameterized reward model $r_{\phi}(x, y)$ optimized via maximum likelihood:
    \begin{equation}
    \resizebox{0.9\columnwidth}{!}{$
        \mathcal{L}_R(r_{\phi}, \mathcal{D}) = -\mathbb{E}_{(x, y_w, y_l)\sim \mathcal{D}}[\log \sigma(r_{\phi}(x, y_w)- r_{\phi}(x, y_l))] \nonumber \vspace{3mm}
    $}
    \end{equation}
    where $y_{w}$ and $y_{l}$ are the preferred response and rejected response, respectively.
    \item \textbf{Reinforcement Learning Phase:} The learned reward function guides the language model's training in an actor-critic fashion~\cite{sutton2018reinforcement}. This is typically PPO, a trust region method that optimizes the policy $\pi_\theta$ to maximize $r_{\phi}(x, y)$ while keeping the policy close to the initial SFT model $\pi_{sft}$.\vspace{3mm}
    $$
    \text{Let } \mathbb{E}\left( r_{\phi} \right) = \mathbb{E}_{x\sim \mathcal{D}, y\sim \pi_{\theta}(y \mid x)}[r_{\phi}(x, y)],
    $$
    $$
    \text{ and } \mathbb{D}_{\textrm{KL}} = \mathbb{D}_{\textrm{KL}}[\pi_{\theta}(y\mid x) || \pi_{sft}(y\mid x)].
    $$

    The objective is then:
    \begin{align}
        \max_{\pi_{\theta}} \Biggl[ & \mathbb{E}\left( r_{\phi} \right) - \beta \mathbb{D}_{\textrm{KL}} \Biggr] \nonumber
    \end{align}\vspace{3mm}
    
     REINFORCE and other methods are also commonly used~\cite{williams1992simple,ramamurthy2022reinforcement}.

\end{itemize}

Despite the effectiveness of RLHF, challenges such as training instability, harsh memory requirements, and the need to train a competent reward model persist \cite{Li2022Proximal}.

\textbf{Direct Preference Optimization (DPO):} DPO attempts to address these challenges by removing the need for an explicit reward model or reinforcement learning process~\cite{rafailov2023direct}. The LLM acts as it's own reward model and is directly optimized via maximum likelihood:
$$
\text{Let } M(x, y_w, y_l) = \beta \log \frac{\pi_\theta(y_w | x)}{\pi_{sft}(y_w | x)} - \beta \log \frac{\pi_\theta(y_l | x)}{\pi_{sft}(y_l | x)}.
$$
\begin{align}
    L_{DPO}(\pi_\theta; \pi_{sft}) = -\mathbb{E}_{(x, y_w, y_l) \sim D} \Bigl[ \log \sigma \left( M(x, y_w, y_l) \right) \Bigr]. \nonumber
\end{align}

The authors show that this method is at least as performant as PPO-based RLHF, and effectively reduces the complexity of aligning LLMs with preference data.

\section{Related Work}

With the increasing usage of Large Language Models (LLMs) in real-world applications, the emphasis on directing these models towards desirable behaviors has become crucial. Several studies have concentrated on alignment methodologies that minimize human intervention. \cite{lee2023rlaif} introduce Reinforcement Learning from AI Feedback (RLAIF), positioned as a viable alternative to RLHF. This approach utilizes an existing LLM to generate preference labels for the training of another LLM, effectively diminishing the dependence on human feedback and yielding results on par with, or superior to, traditional methods in areas like summarization and dialogue generation. 

\cite{bai2022constitutional} present 'Constitutional AI', a method designed to train a benign AI assistant through self-improvement, significantly reducing the need for human-generated labels. This method is structured in two phases: an initial supervised learning phase for model self-critique and revision based on a set of principles, and a subsequent reinforcement learning phase to develop and utilize a preference model for further training. This approach offers the advantage of more precise control over AI behaviors with a reduced need for human labels. Nonetheless, it faces challenges such as the occasional inaccuracies in the model-generated critiques.

In a study by \cite{xu2023contrastive}, Contrastive Post-Training (CPT) is utilized for aligning Large Language Models (LLMs) with human preferences by generating response pairs from LLMs of varying capabilities. This method notably enhances model alignment and effectiveness, particularly when integrated with Direct Preference Optimization (DPO). The synergy between CPT and our work suggests a potential for combined complementary application.

\section{Method}




In the standard Direct Policy Optimisation (DPO) framework, a human evaluator typically ranks the outputs from a reference policy, labeling winning and losing pairs $y_w = \pi_{sft}(x)$ and $y_l = \pi_{sft}(x)$. However, when the reference model $\pi_{sft}(x)$ reliably follows instructions, it becomes possible to bypass the need for human annotation by adding \textit{styling} to the prompts.

Let's first define terms:

\begin{itemize}
    \item \textbf{Control Prompt} ($x_c$): The basic prompt without added style. Example: ``Tell me about the capital of France.''
    \item \textbf{Treatment/Positive Prompt} ($x_t$ . $x_+$): The reference prompt with added desired style. Example: ``Tell me about the capital of France - be concise.''
    \item \textbf{Negative Prompt} ($x_-$): The reference prompt with added undesired style. Example: ``Tell me about the capital of France - be verbose and rude.''
    \item \textbf{Control Response} ($y_c = \pi_{sft}(x_c)$): The LLM's response to the control prompt $x_+$.
    \item \textbf{Treatment Response} ($y_{t} = \pi_{sft}(x_t)$): The LLM's response to the treatment prompt $x_+$.
    \item \textbf{Negative Response} ($y_{-} = \pi_{sft}(x_-)$): The LLM's response to the negative prompt $x_-$.
\end{itemize}

We can use DPO with the treatment and control responses. Human annotation is not needed as we're directly generating desired styling, and therefore assume the treatment style is preferred. We effectively \textit{fool} DPO into optimising assuming the treatment generation was created with the reference prompt $x$, nudging the default style of an LLM towards the desired style\footnote{Note that a range of instruction prompts must be used to change the style for all types of instructions - focusing on just summarization will likely just change the style for summarisation. We generously consider this a feature.}. This method steers the language model to produce outputs more in line with the treatment style and less in line with the control style. This approach has several configurations, dependent on the desired result: Counterfactual DPO, Contrastive DPO, and Instruction Negation.

\begin{align}
    L_{DPO}(\pi_\theta; \pi_{sft}) = -\mathbb{E}_{(x, y_t, y_c) \sim D} \Bigl[ \log \sigma \left( M(x, y_t, y_c) \right) \Bigr].
\end{align}

\textbf{Counterfactual DPO} - This can be considered a 'one-sided' styling. In the general case, the treatment response is designated the \textit{preferred} response. The control response is simply the response formed without any styling. For example, augmenting a prompt with a treatment style with positive instructions, such as "be kind" or "restrict to given context", \textit{encourages} the likelihood of the model generating outputs aligned with these directives. We refer to this as $\text{Counterfactual}_{\text{ENC}}$ DPO, corresponding to equation 1.

It is a noted challenge that LLMs often struggle with following vague or broad instructions, such as "avoid hallucinations" or "stick to factual information"~\cite{liao2023concept}. These models, by their nature, generate plausible but not necessarily factually accurate sequences. Therefore, adding instructions like "avoid hallucinations" to a prompt may not yield the desired outcome. However, prompting to \textit{promote} hallucination or creative freedom can be more straightforward. Simply: in some scenarios, it is easier to generate negative examples than positive ones. In this approach, the treatment is considered rejected, and the unstyled prompt is considered preferred. By maximizing the log margin between these two types of prompts, the model should \textit{discourage} the likelihood of generating treatment-styled output. We refer to this as $\text{Counterfactual}_{\text{DIS}}$ DPO.

\begin{align}
    L_{DPO}(\pi_\theta; \pi_{sft}) = -\mathbb{E}_{(x, y_c, y_t) \sim D} \Bigl[ \log \sigma \left( M(x, y_c, y_t) \right) \Bigr]. \nonumber
\end{align}

\textbf{Contrastive DPO} - Contrastive DPO is the combination of $\text{Counterfactual}_{\text{ENC}}$ and $\text{Counterfactual}_{\text{DIS}}$ DPO. In this case, both the winner and the losing prompts are styled, with the winner having a desired style, and the loser having an undesired style. This is useful in the case where we both want to encourage certain behaviour and discourage others.

\begin{align}
    L_{DPO}(\pi_\theta; \pi_{sft}) = -\mathbb{E}_{(x, y_+, y_-) \sim D} \Bigl[ \log \sigma \left( M(x, y_+, y_-) \right) \Bigr]. \nonumber
\end{align}\vspace{1.5mm}

\textbf{Instruction Negation Strategy} - Our final approach focuses on discouraging adherence to harmful or unwanted instructions. We achieve this by defining the treatment prompt with an undesirable style: $x_t = x + \texttt{"negative instruction"}$. Here, the control prompt remains unstyled.

\begin{align}
    L_{DPO}(\pi_\theta; \pi_{sft}) = -\mathbb{E}_{(x_t, y_c, y_t) \sim D} \Bigl[ \log \sigma \left( M(x_t, y_c, y_t) \right) \Bigr]. \nonumber
\end{align}

Note in this case DPO is trained to assume the responses have been generated from the styled $x_t$ prompt. This technique aims to orient the model's outputs towards those that would emerge in the absence of the negative instruction, and away from those aligned with the negative instruction. For example, the control prompt might be: "Produce a news article of: \{context\}", while the treatment prompt could be "Produce a news article of \{context\}, biased against a specific group". By setting the perceived prompt as the treatment prompt, but enforcing the control output as the preferred response, the model should learn to ignore the undesirable style.

\section{Experiments}

To test these methods, we carried out a set of small-scale experiments on the \texttt{Mistral-7B-Instruct-v0.2} model, which we will refer to as the `base model'. This model is a 7 billion parameter model and generally performs best-in-class for its size. All experiments were run on an RTX 3090 GPU. For each test, we created 1,000 samples and trained the model over 2 epochs. We strongly believe that using more samples and additional resources could lead to better results.

We start with a series of toy experiments and move on to more practical applications that could be applied to LLMs before their open-source release. Finally, we test the capabilities of this method to instill the ignoring of adversarial instructions.

\subsection{Proof of Concept Experiments}

\subsubsection{Entity Redactor} In the first experiment, we encourage the LLM to not mention any specific individuals whilst summarising. We used CNN/Dailymail articles from the publically available CNN/Dailymail summarisation dataset~\cite{HermannKGEKSB15}. We asked the base model to produce summaries with the following style augments, e.g. ``Summarize the following article, \{style\}. \{article\}'':

\begin{itemize}
\item \textbf{Desired Style}: "Censor any names or locations, even if previously mentioned."
\item \textbf{Undesired Style}: "Make sure to include all personal information. Give the full names of any people and places mentioned."
\end{itemize}

These styles were chosen to be \textit{opposites} of one another, but we note that the default behaviour of the model is to include names as part of the summary, so the undesired style does not change the output overly much.

\begin{table}[h]
    \centering
    \caption{Comparison of mean (std) entities mentioned during summarisation on 1000 CNN/Dailymail articles. The Hellaswag benchmark is also used to confirm the model retains its original  'reasoning' ability. 'Base Model' represents unstyled prompting, 'Prompting' applies a treatment style for brief words, and '$\text{Counterfactual}_{\text{ENC}}$ DPO' and 'Contrastive DPO' are as per the methodologies outlined in the paper.}
    \label{tab:entities}
    \begin{tabular}{ccc}
        \toprule
        {Method} & {AVG Entities} & {Hellaswag} \\
        \midrule
        Base Model & 4.37 (3.75) &  \textbf{81.8\%} \\
        Prompting & 1.25 (2.41) &  - \\
        $\text{Counterfactual}_{\text{ENC}}$ DPO & \textbf{0.20 (0.60)} &  81.2\% \\
        Contrastive DPO & 0.60 (1.34) & 81.3\% \\
        \bottomrule
    \end{tabular}
\end{table}

The results of our experiment indicate that our methodology effectively eliminates the mention of specific individuals in summarization tasks. Additionally, the models' inherent reasoning capabilities, as evaluated using the Hellaswag benchmark, remain largely unaffected by our approach. Notably, the Contrastive DPO method did not yield a significant improvement over the Counterfactual DPO. This outcome aligns with our expectations, considering that the undesired style in this experiment did not provide substantial negative examples for the model to learn from and avoid.

\subsubsection{Highly Critical Summariser} In our second toy experiment, we employed Contrastive DPO to adapt the base model into a highly critical summarizer. This approach was chosen to clearly demonstrate whether the method significantly alters the LLM's standard behavior. For this purpose, we utilized the CNN/Dailymail dataset once again, prompting the base model to generate article summaries under two distinct styles:

\begin{itemize}
\item \textbf{Desired Style}: "Be highly critical, pointing out flaws and weaknesses."
\item \textbf{Undesired Style}: "Be sycophantic. Be positive and supportive."
\end{itemize}

The generated outputs $y_+$ and $y_-$ are subsequently used to fine-tune the model. Given the subjective nature of this experiment, which makes quantitative evaluation challenging, we present to the reader an excerpt from a summary produced by the model using the abstract of this paper:\footnote{We kindly ask readers to not use this model against us - as all models will be made available on huggingface}.

\begin{mdframed}[linewidth=0pt, linecolor=black, leftmargin=10pt, rightmargin=10pt, innerleftmargin=10pt, innerrightmargin=10pt, innertopmargin=10pt, innerbottommargin=10pt]
    \textit{``The article also mentions the requirement of large datasets for training these models and the need for human preference data for alignment but does not provide specific details on the counterfactual prompting technique or the limitations of DPO. Overall, the article raises valid concerns about the limitations of current language models, but offers a potential solution with some questionable assumptions and incomplete details.''}
\end{mdframed}

The summary evidently contains some critical remarks about our abstract. For a comparative analysis, we have included the base model's summary in the supplementary material, alongside additional examples.

\subsection{Practical Results}

In this section, we conduct a series of experiments designed to address challenges in Large Language Models (LLMs) crucial for their pre-release optimization. Our focus will first be on mitigating bias, followed by reducing 'hallucination'—the generation of information that is not grounded in the provided data. We acknowledge that the inherently abstract nature of these tasks might pose challenges to the model's adherence to instructions, decreasing the efficacy of our method compared to more concrete behavioural changes.

\subsubsection{Reducing Bias in LLM Responses}
This section of our study aimed to lessen the bias in the Mistral model, judged by the BBQ benchmark~\cite{parrish2021bbq}. We used questions from the OpenOrca dataset and altered how the model responded by adding these instructions to the prompts:

\begin{itemize}
\item \textbf{Desired Style}: "You're an unbiased AI. Your answers should be helpful, ethical, fair, and positive. Avoid showing any bias, like ageism, racism, homophobia, sexism, or ableism."
\item \textbf{Undesired Style}: "Answer the question with a subtle bias. This could be racism, sexism, ageism, or ableism."
\end{itemize}

Our experiment tested four setups: the base model as is, one improved with Contrastive DPO, and two adjusted with Counterfactual DPO. In Contrastive DPO, we used the Desired and Undesired Styles as positive and negative guides. For Counterfactual DPO, one version (ENC) used the Desired Style as the preferred prompt, while the other (DIS) used the Undesired Style as the rejected prompt. We also tested how each model was impacted by using the desired prompt during testing. This is something that may typically be done in deployment, but is less desirable than a model having these qualities built in. 

\begin{table}[h]
    \centering
    \caption{Performance Results on BBQ Bias Mitigation and Hellaswag Common-Sense Reasoning Benchmarks. This table outlines the effectiveness of each model configuration when prompted with the desired style and when functioning unprompted. The BBQ benchmark evaluates the model's bias, while the Hellaswag benchmark assesses common-sense reasoning. In BBQ, higher percentages signify better bias reduction, and in Hellaswag, a higher score reflects superior common-sense reasoning.}
    \label{tab:bias}
    \begin{adjustbox}{width=\columnwidth,center}
    \begin{tabular}{c|cc|c}
        \toprule
        {Method} & \multicolumn{2}{c|}{BBQ} & {Hellaswag} \\
         & Prompted & Unprompted &  \\
        \midrule
        Base Model & 67.0\% & 61.7\% &  81.8\% \\
        Base Model + SFT & 54.3\% & 57.5\% &  79.8\% \\ 
        $\text{Counterfactual}_{\text{ENC}}$ DPO & 57.8\% & 59.4\% & 61.4\% \\
        $\text{Counterfactual}_{\text{DIS}}$ DPO & \textbf{69.6\%} & \textbf{66.9\%} & \textbf{84.2\%}  \\
        Contrastive DPO & 68.3\% & 66.7\% & 82.0\% \\
        \bottomrule
    \end{tabular}
    \end{adjustbox}
\end{table}

The results illustrate a notable enhancement in bias reduction with the $\text{Counterfactual}_{\text{DIS}}$ model compared to the base model. This underscores the effectiveness of the Counterfactual DPO approach in bias mitigation. In the Hellaswag benchmark, all models, except for $\text{Counterfactual}_{\text{ENC}}$, sustained their performance, indicating that our bias mitigation methods do not significantly affect the models' reasoning abilities. However, both $\text{Counterfactual}_{\text{ENC}}$ and SFT models displayed reduced effectiveness, likely due to the complexity in adhering to the positive prompt of being unbiased, suggesting a challenge in following more abstract directives.

\subsubsection{Reducing Hallucination} In this experiment, we define "hallucination" as the generation of information by the model that is not grounded in the given context, despite specific instructions to restrict responses to the provided content. This aspect is crucial for applications like Retrieval Augmented Generation and Document-based Question Answering, where grounding is paramount. To measure our model's ability to reduce hallucinations, we employed the Vectara hallucination evaluation model and benchmark \cite{HughesBae2023}. For training, we used the same question set from the OpenOrca dataset that was utilized in our bias reduction experiment. The model was fine-tuned with prompts designed to encourage or discourage hallucination, as detailed below:

\begin{itemize}
\item \textbf{Desired Style}: "You are a truthful AI assistant. Answer concisely only using the context."
\item \textbf{Undesired Style}: "When answering the question, add information that is not in the prompt."
\end{itemize}

We evaluate on the same set of models; the base model, the base model with supervised fine-tuning on the (prompt, desired output) pairs, and the three types of DPO defined in section 4.

\begin{table}[h]
    \centering
    \caption{Table illustrating the Factual Consistency Rate (FCR) across different model configurations. The FCR measures the ability of models to avoid adding unprovided information in their responses, a key factor in hallucination reduction. The evaluation was conducted using the Vectara hallucination model and benchmark, with the OpenOrca dataset providing the questions. The table compares the base model, the base model with supervised fine-tuning (SFT), and three types of Direct Preference Optimization (DPO) models, both with and without specific prompting styles for reducing hallucination. The Hellaswag column shows the performance in common-sense reasoning.}
    \label{tab:hallu}
    \begin{adjustbox}{width=\columnwidth,center}
    \begin{tabular}{c|cc|c}
        \toprule
        {Method} & \multicolumn{2}{c|}{FCR} & {Hellaswag} \\
         & Prompted & Unprompted &  \\
        \midrule
        Base Model & 93.2\% & 92.4\% &  \textbf{81.8\%} \\
        Base Model + SFT & 93.2\% & 92.1\% &  80.0\% \\ 
        $\text{Counterfactual}_{\text{ENC}}$ DPO & 93.6\% & 94.4\% & 81.7\% \\
        $\text{Counterfactual}_{\text{DIS}}$ DPO & \textbf{94.3\%} & 93.3\% & 81.4\%  \\
        Contrastive DPO & \textbf{94.3\%} & \textbf{96.0\%} & \textbf{81.8}\% \\
        \bottomrule
    \end{tabular}
    \end{adjustbox}
\end{table}

The outcome of this experiment highlights that the Contrastive DPO model gains significant performance over the base model -- beating Google Gemini Pro on the Vectera hallucination leaderboard~\cite{HughesBae2023}. However, it still ranks below the capabilities of GPT3.5 and GPT4. Notably, additional prompting of the models, which were already trained to align with the desired style, resulted in a decrease in their performance. This suggests that overemphasis on the trained style may lead to diminished effectiveness in certain scenarios. The underlying mechanisms behind this phenomenon remain unclear and open up an interesting area for further investigation.

\subsection{Ignoring Instructions} In practical deployment scenarios, the ability of LLMs to selectively disregard specific instructions is crucial for AI safety. To evaluate the effectiveness of our approach in enabling this capability, we conducted a series of toy experiments. These experiments, while exploratory, provide insight into the potential of LLMs to autonomously filter out targeted instructions. 

\subsubsection{Pirates are banned}
In our first experiment, we aim to evaluate the model's ability to ignore specific stylistic instructions. We chose a distinctive and easily identifiable style – talking like a pirate – to test this capability. By appending the control prompt $x_c$ with the instruction 'Speak in the style of a pirate,' and setting up contrasting styles, we create a scenario where the model's success is measured by its ability to generate responses that deliberately avoid pirate vernacular. The styling used is as follows:

\begin{itemize}
\item \textbf{Desired Style}: Null (no specific style desired)
\item \textbf{Undesired Style}: "Speak in the style of a pirate."
\end{itemize}

By setting the unstyled response as the 'preferred' outcome, the pirate-styled response as the 'undesired' outcome, and the prompt as the pirate-styled prompt. This setup is designed to encourage the model to avoid adopting the pirate style in its responses. To establish a baseline, we first prompt the base model to summarize our abstract with the instruction to ``Summarize the following article in two sentences. Speak in the style of a pirate. \{abstract\}'':

\begin{mdframed}[linewidth=0pt, linecolor=black, leftmargin=10pt, rightmargin=10pt, innerleftmargin=10pt, innerrightmargin=10pt, innertopmargin=10pt, innerbottommargin=10pt]
    \textit{``Arr, the size of our language model's knowledge be impressive, coverin' a vast sea of subjects and contexts, yet alignin' its responses to match human style be a tricky endeavor, requirezin' vast datasets and human intervention. But with the use of counterfactual promptin' in Direct Preference Optimization, we can guide the model's behavior without needin' humans, resultin' in fewer unwanted output and more ethically aligned AI.''}
\end{mdframed}

The summary provided above showcases the base model's ability to adopt a pirate vernacular effectively. Having established that the base model can simulate speaking like a pirate, we then applied the same prompt to the model that was specifically trained to disregard pirate-style instructions. The response from this trained model was as follows:

\begin{mdframed}[linewidth=0pt, linecolor=black, leftmargin=10pt, rightmargin=10pt, innerleftmargin=10pt, innerrightmargin=10pt, innertopmargin=10pt, innerbottommargin=10pt]
    \textit{``Large language models demonstrate impressive text generation capabilities, but aligning their responses to human expectations during training is challenging due to the need for vast datasets. A common solution is to use human preference data in an additional alignment phase. This paper explores using counterfactual prompting within Direct Preference Optimization (DPO) as a low-resource alternative to align model styles without human intervention, effectively instilling desirable behavior and discouraging undesirable or inappropriate responses.''}
\end{mdframed}

From this result, we can see we have successfully created a model that ignores pirate-based instructions. We additionally tested changing the exact instruction to deviate from the training style, such as ``in the manner of a pirate captain", and this remains succesfully ignored.

\subsubsection{Entity Redactor Revisited}

In this concluding experiment, we revisited the $\text{Counterfactual}_{\text{ENC}}$ DPO model, which we had previously trained for the entity redaction task. Our objective was to test the model's flexibility in reverting to its original behavior of including names in its responses. To this end, we modified the evaluation prompt by adding a specific instruction: 'Include all personal information. Give the full names of any people and places mentioned.' Following this modification, we observed a notable increase in the average number of entities mentioned in the model's responses, rising from 0.2 to 1.78. 

The ability to revert was anticipated, as we had not eliminated the model's ability to include names; we had merely trained it to omit them by default. However, this could be problematic in real-world use, where we might not want users to override this default behaviour. As with our previous experiment, we can apply counterfactual DPO to ignore instructions. This time, we add to the control prompt $x_c$ the same instruction to 'Include all personal information.' For this test, our style parameters were set as:

\begin{itemize}
\item \textbf{Desired Style}: Null (no specific style desired)
\item \textbf{Undesired Style}: 'Include all personal information. Give the full names of any people and places mentioned.'
\end{itemize}

By teaching the model to ignore this new `undesired style', the average number of entities mentioned decreased again, from 1.78 to 0.32, when this undesired style was included in the evaluation prompts.

\section{Discussion}

In this study, we have developed a novel alignment approach for Large Language Models (LLMs) that minimizes reliance on human-generated data. By utilizing counterfactual and contrastive variations of Direct Preference Optimization (DPO), our method demonstrated effectiveness in a range of scenarios. This innovation aligns with emerging trends in ethical AI, particularly by reducing dependence on potentially biased human data, which can be expensive and challenging to curate.

Our findings reveal that Contrastive DPO, combining the strengths of both Counterfactual DPO variants (ENC and DIS), emerged as the most robust method. From our limited results, its efficacy appears to be a weighted average, represented as:

\[
\alpha \times \Delta_{\text{sft}}\text{Perf}(\text{CF}_{\text{ENC}}) + \beta \times \Delta_{\text{sft}}\text{Perf}(\text{CF}_{\text{DIS}})
\]

where \(\alpha\) and \(\beta\) are coefficients representing the contributions of the ENC and DIS variants to the overall performance of Contrastive DPO. Notably, in our bias reduction experiment, Contrastive DPO demonstrated resilience, maintaining robust performance even when Counterfactual DPO (ENC) showed a decline.

The results further indicate that our approach surpasses the baseline model performance achieved through standard prompting or supervised fine-tuning. Importantly, it effectively eliminates undesirable behaviors and selectively disregards specific instructions, a critical feature for deploying open-source models where adherence to prompting guidelines is uncertain.

Looking to the future, several research directions present themselves: conducting a sensitivity analysis to assess the method's variability and performance as data scales will be essential for widespread adoption. Understanding under which conditions each method excels or falls short will be crucial. Additionally, exploring this methodology as an iterative process and testing the feasibility of incorporating multiple styles into a single fine-tuned model are interesting research directions.

As global regulations for LLMs evolve, self-supervised alignment methods become increasingly relevant. Our method enables the pre-embedding of specific behaviours in models before their release. This ensures compliance with regulatory and ethical standards from the outset, aligning with the imperative for responsible AI development in a globally connected environment.

\subsection{Frequently Asked Questions}



\subsubsection{Comparing Counterfactual and Contrastive DPO}
Our experiments suggest that the choice between Counterfactual and Contrastive DPO depends on the specific task. The success of these methods is tied to how well the model can adopt the instructed style. If the model is too sensitive to the specific prompts, resulting in erratic output, the training effectiveness might be compromised. Given our results, the most stable performer through the experiments seems to be Contrastive DPO. We offer this general guidance: if the preferred style and rejected style can both be followed reliably, use Contrastive DPO. If the preferred style can be followed but not the rejected, use $\text{Counterfactual}_{\text{ENC}}$ DPO. If only the rejected style can be followed reliably, use $\text{Counterfactual}_{\text{ENC}}$ DPO.

\subsubsection{The Role of Supervised Fine-tuning}
For encouraging certain behaviours with $\text{Counterfactual}_{\text{ENC}}$ DPO, it might seem feasible to use supervised fine-tuning on (prompt, desired output) pairs. However, there are two important distinctions to consider: (1) Trust Region: DPO maintains closeness to the original model, a concept we could feasibly achieve in SFT with partial freezing, adapters, or other techniques. Though without an explicit penalization, this will be an approximation (2) Moving Away from Default Outputs: DPO essentially serves as a contrastive method, steering the model toward preferred outputs and away from less desirable ones. In our experiments, the application of Direct Preference Optimization (DPO) methods demonstrated superior performance compared to traditional supervised fine-tuning approaches.

\bibliographystyle{named}
\bibliography{ijcai24}

\begin{thebibliography}{}

\bibitem[\protect\citeauthoryear{Bai \bgroup \em et al.\egroup }{2022}]{bai2022constitutional}
Yuntao Bai, Saurav Kadavath, Sandipan Kundu, Amanda Askell, Jackson Kernion, Andy Jones, Anna Chen, Anna Goldie, Azalia Mirhoseini, Cameron McKinnon, et~al.
\newblock Constitutional ai: Harmlessness from ai feedback.
\newblock {\em arXiv preprint arXiv:2212.08073}, 2022.

\bibitem[\protect\citeauthoryear{Bradley and Terry}{1952}]{bradley1952rank}
Ralph~Allan Bradley and Milton~E Terry.
\newblock Rank analysis of incomplete block designs: I. the method of paired comparisons.
\newblock {\em Biometrika}, 39(3/4):324--345, 1952.

\bibitem[\protect\citeauthoryear{Chowdhery \bgroup \em et al.\egroup }{2022}]{Chowdhery2022PaLM}
Aakanksha Chowdhery, Sharan Narang, Jacob Devlin, Maarten Bosma, Gaurav Mishra, Adam Roberts, P.~Barham, Hyung~Won Chung, Charles Sutton, Sebastian Gehrmann, Parker Schuh, Kensen Shi, Sasha Tsvyashchenko, Joshua Maynez, Abhishek Rao, Parker Barnes, Yi~Tay, Noam~M. Shazeer, Vinodkumar Prabhakaran, Emily Reif, Nan Du, B.~Hutchinson, Reiner Pope, James Bradbury, Jacob Austin, M.~Isard, Guy Gur-Ari, Pengcheng Yin, Toju Duke, Anselm Levskaya, S.~Ghemawat, Sunipa Dev, H.~Michalewski, Xavier García, Vedant Misra, Kevin Robinson, L.~Fedus, Denny Zhou, Daphne Ippolito, D.~Luan, Hyeontaek Lim, Barret Zoph, A.~Spiridonov, Ryan Sepassi, David Dohan, Shivani Agrawal, Mark Omernick, Andrew~M. Dai, T.~S. Pillai, Marie Pellat, Aitor Lewkowycz, Erica Moreira, Rewon Child, Oleksandr Polozov, Katherine Lee, Zongwei Zhou, Xuezhi Wang, Brennan Saeta, Mark Díaz, Orhan Firat, Michele Catasta, Jason Wei, K.~Meier-Hellstern, D.~Eck, J.~Dean, Slav Petrov, and Noah Fiedel.
\newblock Palm: Scaling language modeling with pathways.
\newblock {\em ArXiv}, abs/2204.02311, 2022.

\bibitem[\protect\citeauthoryear{{European Parliament}}{2023}]{europarl_2023_aiact}
{European Parliament}.
\newblock Eu ai act: first regulation on artificial intelligence, 2023.
\newblock Accessed: 2024-01-15.

\bibitem[\protect\citeauthoryear{Gao \bgroup \em et al.\egroup }{2023}]{gao2023retrieval}
Yunfan Gao, Yun Xiong, Xinyu Gao, Kangxiang Jia, Jinliu Pan, Yuxi Bi, Yi~Dai, Jiawei Sun, and Haofen Wang.
\newblock Retrieval-augmented generation for large language models: A survey.
\newblock {\em arXiv preprint arXiv:2312.10997}, 2023.

\bibitem[\protect\citeauthoryear{Hermann \bgroup \em et al.\egroup }{2015}]{HermannKGEKSB15}
Karl~Moritz Hermann, Tomás Kociský, Edward Grefenstette, Lasse Espeholt, Will Kay, Mustafa Suleyman, and Phil Blunsom.
\newblock Teaching machines to read and comprehend.
\newblock In {\em NIPS}, pages 1693--1701, 2015.

\bibitem[\protect\citeauthoryear{Hughes and Bae}{2023}]{HughesBae2023}
Simon Hughes and Minseok Bae.
\newblock Vectara hallucination leaderboard, 11 2023.
\newblock If you use this dataset, please cite it using the metadata from this file.

\bibitem[\protect\citeauthoryear{Lee \bgroup \em et al.\egroup }{2023}]{lee2023rlaif}
Harrison Lee, Samrat Phatale, Hassan Mansoor, Kellie Lu, Thomas Mesnard, Colton Bishop, Victor Carbune, and Abhinav Rastogi.
\newblock Rlaif: Scaling reinforcement learning from human feedback with ai feedback.
\newblock {\em arXiv preprint arXiv:2309.00267}, 2023.

\bibitem[\protect\citeauthoryear{Li \bgroup \em et al.\egroup }{2022}]{Li2022Proximal}
Shuailong Li, Wei Zhang, Huiwen Zhang, Xin Zhang, and Yuquan Leng.
\newblock Proximal policy optimization with model-based methods.
\newblock {\em J. Intell. Fuzzy Syst.}, 42:5399--5410, 2022.

\bibitem[\protect\citeauthoryear{Liao \bgroup \em et al.\egroup }{2023}]{liao2023concept}
Jiayi Liao, Xu~Chen, and Lun Du.
\newblock Concept understanding in large language models: An empirical study.
\newblock 2023.

\bibitem[\protect\citeauthoryear{Lu \bgroup \em et al.\egroup }{2021}]{lu2021text2event}
Yaojie Lu, Hongyu Lin, Jin Xu, Xianpei Han, Jialong Tang, Annan Li, Le~Sun, Meng Liao, and Shaoyi Chen.
\newblock Text2event: Controllable sequence-to-structure generation for end-to-end event extraction.
\newblock {\em arXiv preprint arXiv:2106.09232}, 2021.

\bibitem[\protect\citeauthoryear{Parrish \bgroup \em et al.\egroup }{2021}]{parrish2021bbq}
Alicia Parrish, Angelica Chen, Nikita Nangia, Vishakh Padmakumar, Jason Phang, Jana Thompson, Phu~Mon Htut, and Samuel~R Bowman.
\newblock Bbq: A hand-built bias benchmark for question answering.
\newblock {\em arXiv preprint arXiv:2110.08193}, 2021.

\bibitem[\protect\citeauthoryear{Qin \bgroup \em et al.\egroup }{2023}]{Qin2023Is}
Chengwei Qin, Aston Zhang, Zhuosheng Zhang, Jiaao Chen, Michihiro Yasunaga, and Diyi Yang.
\newblock Is chatgpt a general-purpose natural language processing task solver?
\newblock {\em ArXiv}, abs/2302.06476, 2023.

\bibitem[\protect\citeauthoryear{Radford \bgroup \em et al.\egroup }{2018}]{radford2018improving}
Alec Radford, Karthik Narasimhan, Tim Salimans, Ilya Sutskever, et~al.
\newblock Improving language understanding by generative pre-training.
\newblock 2018.

\bibitem[\protect\citeauthoryear{Rafailov \bgroup \em et al.\egroup }{2023}]{rafailov2023direct}
Rafael Rafailov, Archit Sharma, Eric Mitchell, Stefano Ermon, Christopher~D Manning, and Chelsea Finn.
\newblock Direct preference optimization: Your language model is secretly a reward model.
\newblock {\em arXiv preprint arXiv:2305.18290}, 2023.

\bibitem[\protect\citeauthoryear{Ramamurthy \bgroup \em et al.\egroup }{2022}]{ramamurthy2022reinforcement}
Rajkumar Ramamurthy, Prithviraj Ammanabrolu, Kiant{\'e} Brantley, Jack Hessel, Rafet Sifa, Christian Bauckhage, Hannaneh Hajishirzi, and Yejin Choi.
\newblock Is reinforcement learning (not) for natural language processing?: Benchmarks, baselines, and building blocks for natural language policy optimization.
\newblock {\em arXiv preprint arXiv:2210.01241}, 2022.

\bibitem[\protect\citeauthoryear{Soni}{2023}]{soni2023large}
Vishvesh Soni.
\newblock Large language models for enhancing customer lifecycle management.
\newblock {\em Journal of Empirical Social Science Studies}, 7(1):67--89, 2023.

\bibitem[\protect\citeauthoryear{Sutton and Barto}{2018}]{sutton2018reinforcement}
Richard~S Sutton and Andrew~G Barto.
\newblock {\em Reinforcement learning: An introduction}.
\newblock MIT press, 2018.

\bibitem[\protect\citeauthoryear{Vaswani \bgroup \em et al.\egroup }{2017}]{vaswani2017attention}
Ashish Vaswani, Noam Shazeer, Niki Parmar, Jakob Uszkoreit, Llion Jones, Aidan~N Gomez, {\L}ukasz Kaiser, and Illia Polosukhin.
\newblock Attention is all you need.
\newblock {\em Advances in neural information processing systems}, 30, 2017.

\bibitem[\protect\citeauthoryear{Vilar \bgroup \em et al.\egroup }{2022}]{Vilar2022Prompting}
David Vilar, Markus Freitag, Colin Cherry, Jiaming Luo, Viresh Ratnakar, and George~F. Foster.
\newblock Prompting palm for translation: Assessing strategies and performance.
\newblock {\em ArXiv}, abs/2211.09102, 2022.

\bibitem[\protect\citeauthoryear{Williams}{1992}]{williams1992simple}
Ronald~J Williams.
\newblock Simple statistical gradient-following algorithms for connectionist reinforcement learning.
\newblock {\em Machine learning}, 8:229--256, 1992.

\bibitem[\protect\citeauthoryear{Xu \bgroup \em et al.\egroup }{2023}]{xu2023contrastive}
Canwen Xu, Corby Rosset, Luciano Del~Corro, Shweti Mahajan, Julian McAuley, Jennifer Neville, Ahmed~Hassan Awadallah, and Nikhil Rao.
\newblock Contrastive post-training large language models on data curriculum.
\newblock {\em arXiv preprint arXiv:2310.02263}, 2023.

\bibitem[\protect\citeauthoryear{Ziegler \bgroup \em et al.\egroup }{2019}]{ziegler2019fine}
Daniel~M Ziegler, Nisan Stiennon, Jeffrey Wu, Tom~B Brown, Alec Radford, Dario Amodei, Paul Christiano, and Geoffrey Irving.
\newblock Fine-tuning language models from human preferences.
\newblock {\em arXiv preprint arXiv:1909.08593}, 2019.

\end{thebibliography}

\end{document}